\documentclass{article}

\PassOptionsToPackage{numbers,sort}{natbib}
\usepackage[final]{neurips_2018}

\usepackage[utf8]{inputenc} % allow utf-8 input
\usepackage[T1]{fontenc}    % use 8-bit T1 fonts
\usepackage{hyperref}       % hyperlinks
\usepackage{graphicx}
\usepackage{siunitx}
\usepackage{url}            % simple URL typesetting
\usepackage{booktabs}       % professional-quality tables
\usepackage{amsmath}
\usepackage{amsfonts}       % blackboard math symbols
\usepackage{nicefrac}       % compact symbols for 1/2, etc.
\usepackage{microtype}      % microtypography
\usepackage{color}
\definecolor{mydarkblue}{rgb}{0,0.08,0.45}
\hypersetup{
	colorlinks=true,
	linkcolor=mydarkblue,
	citecolor=mydarkblue,
	filecolor=mydarkblue,
	urlcolor=mydarkblue,
}

\title{On the Inductive Bias of Word-Character-Level Multi-Task Learning for Speech Recognition}

\author{
	Jan Kremer \\
	Corti, Copenhagen, Denmark \\
	\texttt{jk@corti.ai} \\
	\And
	Lasse Borgholt \\
	Corti, Copenhagen, Denmark \\
	\texttt{lb@corti.ai} \\
	\And
	Lars Maaløe \\
	Corti, Copenhagen, Denmark \\
	\texttt{lm@corti.ai}
}

\begin{document}
	
	\maketitle
	
	\begin{abstract}
		End-to-end automatic speech recognition (ASR) commonly transcribes audio signals into sequences of characters while its performance is evaluated by measuring the word-error rate (WER). This suggests that predicting sequences of words directly may be helpful instead. However, training with word-level supervision can be more difficult due to the sparsity of examples per label class. In this paper we analyze an end-to-end ASR model that combines a word-and-character representation in a multi-task learning (MTL) framework. We show that it improves on the WER and study how the word-level model can benefit from character-level supervision by analyzing the learned inductive preference bias of each model component empirically. We find that by adding character-level supervision, the MTL model interpolates between recognizing more frequent words (preferred by the word-level model) and shorter words (preferred by the character-level model).
	\end{abstract}
	
	\section{Introduction}
	End-to-end automatic speech recognition (ASR) allows for learning a direct mapping from audio signals to character outputs. 
	Usually, a language model re-scores the predicted transcripts during inference to correct spelling mistakes \citep{hannun_deep_2014}. 
	If we map the audio input directly to words, we can use a simpler decoding mechanism and reduce the prediction time. 
	Unfortunately, word-level models can only be trained on known words. 
	Out-of-vocabulary (OOV) words have to be mapped to an unknown token. 
	Furthermore, decomposing transcripts into sequences of words decreases the available number of examples per label class. 
	These shortcomings make it difficult to train on the word-level \citep{audhkhasi_direct_2017}. 
	
	Recent works have shown that multi-task learning (MTL) \citep{caruana_multitask_1997} on the word- and character-level can improve the word-error rate (WER) of common end-to-end speech recognition architectures \citep{kim_joint_2017,audhkhasi_direct_2017,li_acoustic--word_2017,audhkhasi_building_2018,li_advancing_2018,sanabria_hierarchical_2018,ueno_acoustic--word_2018}. MTL can be interpreted as learning an inductive bias with favorable generalization properties \citep{baxter_model_2000}. In this work we aim at characterizing the nature of this inductive bias in word-character-level MTL models by analyzing the distribution of words that they recognize. Thereby, we seek to shed light on the learning process and possibly inform the design of better models. 
	We will focus on connectionist temporal classification (CTC) \citep{graves_connectionist_2006}. However, the analysis can also prove beneficial to other modeling paradigms, such as RNN Transducers \citep{graves_sequence_2012} or Encoder-Decoder models, e.g.,   \citep{bahdanau_end--end_2016,chan_listen_2016}.
	\paragraph{Contributions.} 
	We show that, contrary to earlier negative results \citep{audhkhasi_direct_2017,soltau_neural_2017}, it is in fact possible to train a word-level model from scratch on a relatively small dataset and that its performance can be further improved by adding character-level supervision. 
	Through an empirical analysis we show that the resulting MTL model combines the preference biases of word- and character-level models. 
	We hypothesize that this can partially explain why word-character MTL improves on only using a single decomposition, such as phonemes, characters or words.
	
	\section{Related work}
	Several works have explored using words instead of characters or phonemes as outputs of the end-to-end ASR model \citep{soltau_neural_2017,audhkhasi_direct_2017}. \citet{soltau_neural_2017} found that in order to solve the problem of observing only few labels per word, they needed to use a large dataset of $120,000$ hours to train a word-level model directly. Accordingly, \citet{audhkhasi_direct_2017} reported difficulty to train a model on words from scratch and instead fine-tuned a pre-trained character-level model after replacing the last dense layer with a word embedding. 
	
	MTL enables a straightforward joint training procedure to integrate transcript information on multiple levels of granularity. Treating word- and character-level transcription as two distinct tasks allows for combining their losses in a parallel \citep{li_acoustic--word_2017,toshniwal_multitask_2017,ueno_acoustic--word_2018,li_advancing_2018} or hierarchical structure \citep{fernandez_sequence_2007,krishna_hierarchical_2018,sanabria_hierarchical_2018}. Augmenting the commonly-used CTC loss with an attention mechanism can help with aligning the predictions on both character- and word-level \citep{audhkhasi_building_2018,das_advancing_2018,li_advancing_2018}.
	All these MTL methods improve a standard CTC baseline.
	
	Finding the right granularity of the word decomposition is in itself a difficult problem. While \citet{li_advancing_2018} used different fixed decompositions of words, sub-words and characters,  it is also possible to optimize over alignments and decompositions jointly \citep{liu_gram-ctc_2017}. Orthogonal to these works different authors have explored how to minimize WER directly by computing approximate gradients \citep{shannon_optimizing_2017,zhou_improving_2018}.
	
	When and why does MTL work? Earlier theoretical work argued that the auxiliary task provides a favorable inductive bias to the main task \citep{baxter_model_2000}. Within natural language processing on text several works verified empirically that this inductive bias is favorable if there is a certain notion of relatedness between the tasks \citep{sogaard_deep_2016,bingel_identifying_2017,augenstein_multi-task_2018}. Here, we investigate how to characterize the inductive bias learned via MTL for speech recognition.
	\section{Combining word- and character-level ASR}
	The CTC loss is defined as follows \citep{graves_connectionist_2006}:
	\begin{equation*}
	\mathcal{L}(\mathbf{x}, \mathbf{z}) := - \log \sum_{\boldsymbol{\pi} \in \mathcal{B}^{-1}(\mathbf{z})} p(\boldsymbol{\pi}|\mathbf{x}) = - \log \sum_{\boldsymbol{\pi} \in \mathcal{B}^{-1}(\mathbf{z})}  \prod_t p(\pi_t|\mathbf{x}) \enspace,
	\end{equation*}
	where $\mathbf{x}$ is the audio input, commonly a spectrogram, and $\boldsymbol{\pi}$ is a path that corresponds to the ground-truth transcript $\mathbf{z}$. The squashing function $\mathcal{B}$ maps a path $\boldsymbol{\pi}$ to the output $\mathbf{z}$ by first merging repetitions and then deleting so-called blank tokens. The gradient of the CTC loss can be computed efficiently using a modified forward-backward algorithm.
	
	Typically, $\pi_t$ is a categorical random variable over the corresponding output alphabet $\mathcal{A} = \{a, b, c, ..., \epsilon \} $. Here, $\epsilon$ is the blank token which encodes the empty string. This output representation enables the model to be able to transcribe any word possible without a specified alignment.
	Character-level CTC models are often supplemented by an external language model that can significantly improve the accuracy of the ASR. This is because these models still make spelling mistakes despite being trained on large amounts of data \citep{amodei_deep_2016}.
	
	By using an alphabet of words one can ensure that there are no misspellings. The alphabet could contain, for example, the most common words found in the training set. This has the advantage that any word is guaranteed to be spelled correctly and that costly re-scoring on a character-level is avoided. However, by using a word-level decoding, we can no longer predict rare or new words. In this case the model has to be content with outputting an unknown token. Another challenge when using a word-level model is label sparsity. While we will observe many examples of a single character, there will be fewer for a single word, making overfitting more likely.  We aim at counter-acting these shortcomings by making use of character-level information during training, similar to \citet{audhkhasi_direct_2017}.
	
	In this work we combine word- and character-level models via an MTL loss and denote this a word-character-level model. We treat each output-level prediction as a separate task and form a linear combination of the losses. The MTL loss is then defined as 
	\begin{equation*}
	\mathcal{L}_\text{MTL}(\mathbf{x}, \mathbf{z}) := \mathcal{L}_\text{word}(\mathbf{x}, \mathbf{z}) + \lambda \mathcal{L}_\text{char}(\mathbf{x}, \mathbf{z})\enspace,
	\end{equation*}
	where $\lambda \geq 0$ defines a hyperparameter to weight the influence of the character-level CTC loss $\mathcal{L}_\text{char}$ against the word-level CTC loss $\mathcal{L}_\text{word}$. In our experiments we set it to $1$, giving equal contribution to both loss terms, but other choices may improve the performance. Alternatively, one could try to estimate this  weight  based on the uncertainty \citep{kendall_multi-task_2018} or gradient norm \citep{chen_gradnorm_2018} of each loss term. We experimented with these approaches, but did not observe any significant improvement in performance over the equally-weighted loss.
	\section{Experiments}
	We trained our models using a convolutional architecture which is based on Wav2Letter \citep{collobert_wav2letter_2016}. 
	Details can be found in the appendix. 
	Compared to recurrent neural networks, convolutional neural networks avoid iterative computation over time and suffer less from the vanishing/exploding gradient problem. 
	They achieve comparable performance in terms of WER \citep{collobert_wav2letter_2016,zhang_towards_2016}.
	We performed all experiments on read news articles from the Wall Street Journal (WSJ) \citep{woodland_large_1994}. 
	This dataset has relatively little background noise and allows us to focus on the influence of word frequency and word length. 
	We used the \emph{si284} subset for training, and \emph{dev92} for validation. 
	For the character-level model we used $32$ different characters which include the space-character and a blank token. 
	To define the output alphabet for the word-level model, we included all words that appeared at least $5$ times in the training set in addition to a blank and an unknown token. 
	This corresponds to an alphabet of $9411$ units with an OOV rate of $\SI{9}{\percent}$ on the training set, and $\SI{10}{\percent}$ on the validation set, which represents a lower bound for the achievable WER of a word-level model. 
	For the MTL model we let word- and character-level model share every layer but the last.
	
	To decode the output on the character- and word-level, we used greedy decoding. In order to get rid of unknown tokens in our prediction, we employed the following heuristic \citep{li_acoustic--word_2017}: For each unknown token predicted on the word-level, we substituted the corresponding word on the character-level that was defined at the same time step. To compare our results we also trained word- and character-only models. For optimization we used the Adam-optimizer \citep{kingma_adam_2015} with a learning rate of ${5}\mathrm{e}{-4}$ and a batch size of $16$ to fit the whole model into the memory of one GPU. We applied batch normalization and dropout. For the input data, we transformed each utterance into spectrograms over $\SI{20}{\ms}$ with a shift of $\SI{10}{\ms}$ using $40$ log-mel coefficients, standardized per spectrogram. We ran each experiment for $100$ epochs, corresponding to $233,838$ updates.
	
\begin{figure}[t]
		\centering
		\includegraphics[width=0.59\linewidth]{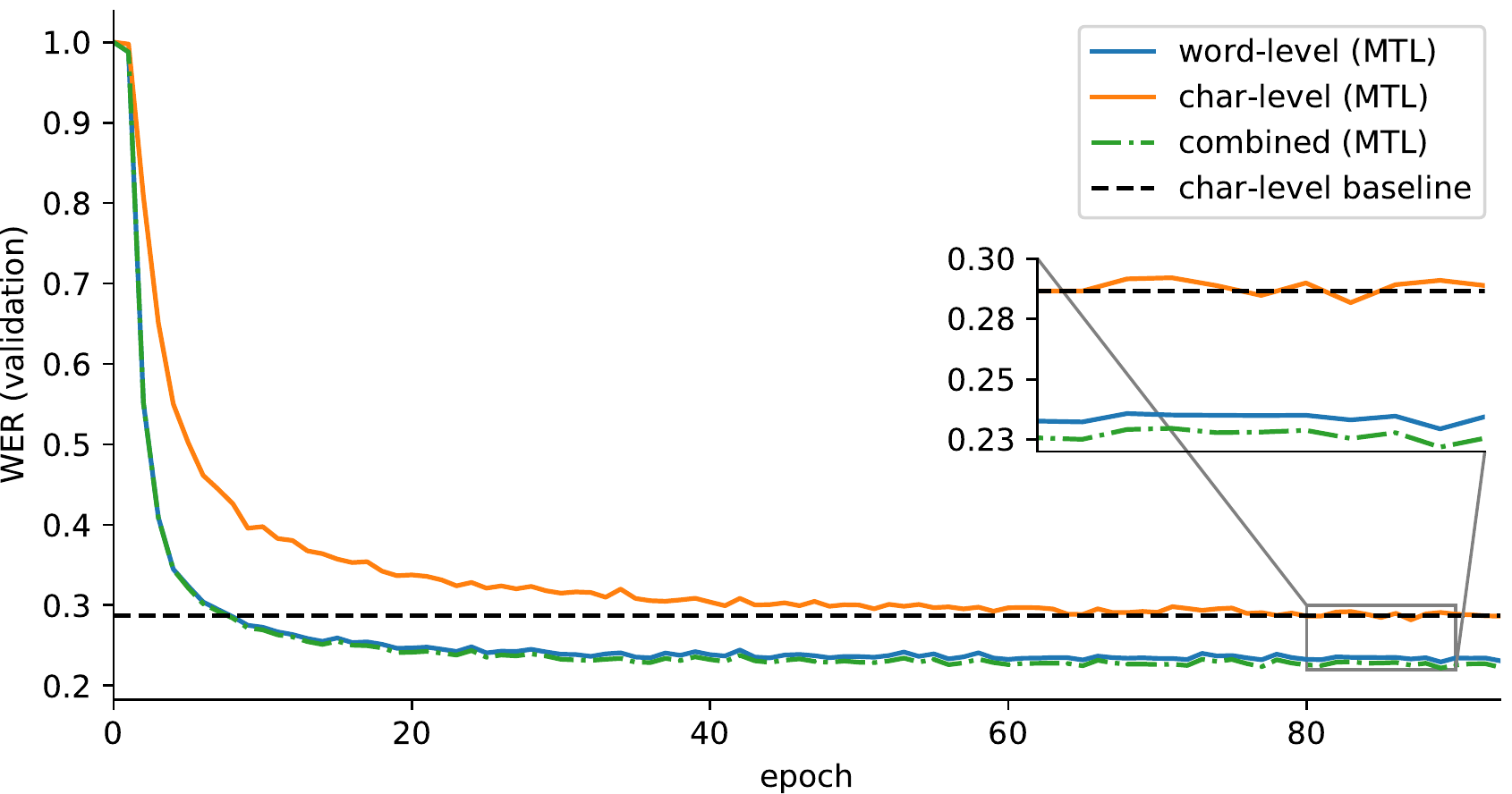}
		\caption{Multi-task learning with equal weighting on word- and character-level.}
		\label{fig:mtl_equal}
	\end{figure}

	\paragraph{MTL performance.}
	The results of our experiments can be found in Figure~\ref{fig:mtl_equal}. 
	It shows the learning curve for the word- and character-level components by measuring the WER on the validation set. 
	The dashed line shows the achieved WER using a character-level model without joint word-level training. We observe that MTL converges faster and to a lower WER of $\SI{23}{\percent}$, which is $5$ percentage points lower than the character-level component of the MTL network, or the single-task character-level baseline. 
	Using a beam search decoder with a lexicon constraint on the character-level model reduces the WER from $\SI{28}{\percent}$ to a WER of $\SI{24}{\percent}$, which is still higher than our MTL error. This shows that MTL performs favorably even without a language model. 
	A word-level-only model achieved the same performance as the character-level baseline on this dataset. 
	Contrary to the findings of \citet{audhkhasi_direct_2017}, this shows that it is indeed possible to train a word-level model from scratch, even without a large amount of training data.
	While the combined decoding only gives an improvement of $0.7$ percentage points in terms of WER, it eliminates unknown-token predictions which might make transcripts more readable.
	\paragraph{Characterizing the inductive bias.}
	\citet{arpit_closer_2017} have shown that a neural network trained with stochastic gradient descent learns easier examples first. 
	We argue that we can characterize the preference bias of our model and learning algorithm by showing which examples are easy to classify in the particular representation that each of the models is learning. 
	Since ASR models are usually evaluated in terms of WER, we consider which words each model is learning. 
	To this end we chose a relatively clean dataset and considered the attributes frequency and length to describe a word. 
	
\begin{figure}[t]
		\centering
		\includegraphics[width=0.8\linewidth]{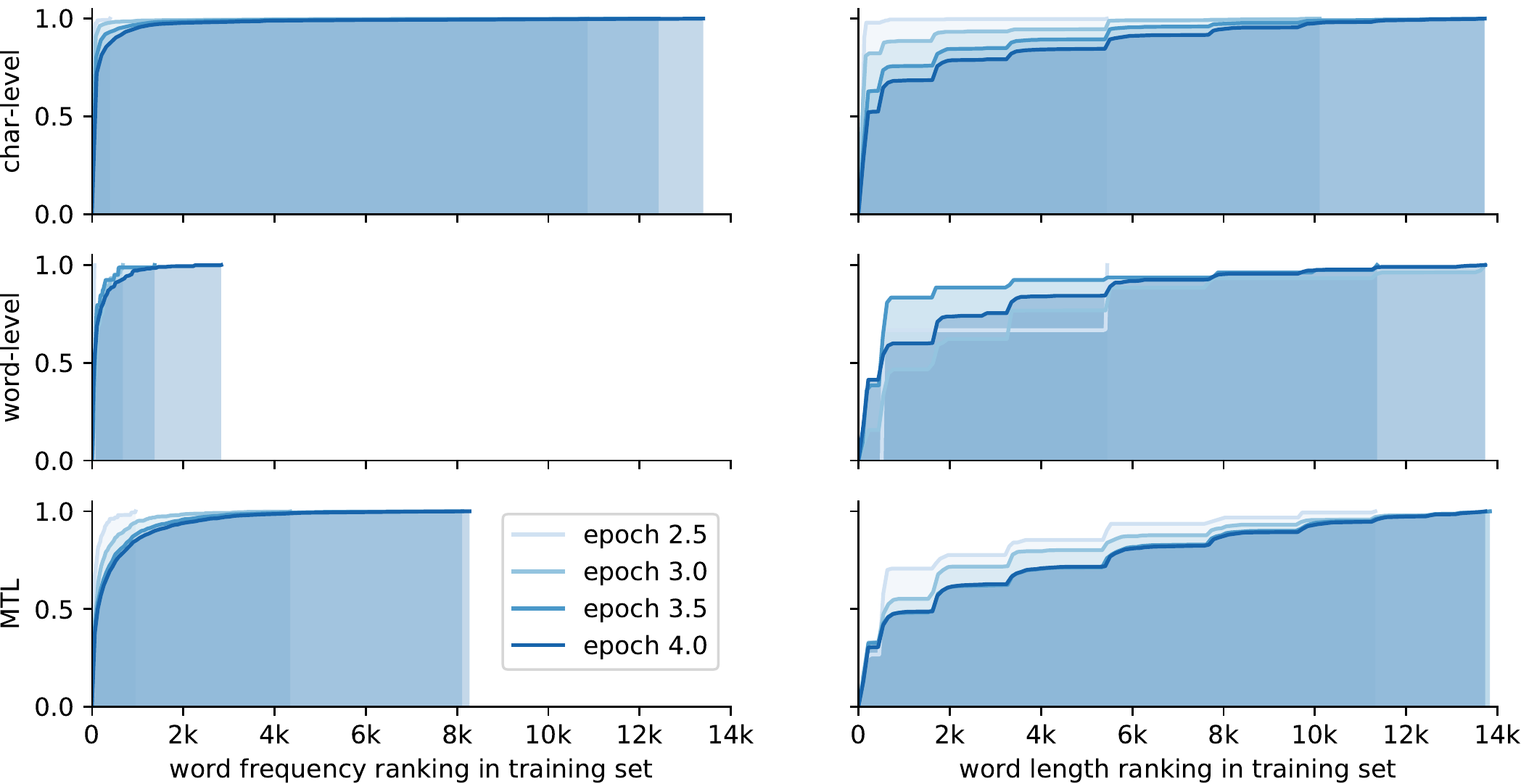}
		\caption{Comparing the CDFs of recognized words between character-, word- and MTL-model. ({\bf left}) The distribution of recognized words during the first epochs as a function of word frequency rank. The most frequent words in the training set are on the left. ({\bf right}) The distribution of recognized words during the first epochs as a function of word length rank. Longer words are on the right.}
		\label{fig:words}
	\end{figure}

	We trained each model for $4$ epochs and recorded the distribution of the recognized words during training. 
	Since we are not given a perfect alignment between speech and ground-truth transcript, we define a word as being recognized if it is both present in the greedy prediction on the validation set and the corresponding ground-truth transcript. 
	Figure~\ref{fig:words} shows how the distribution of recognized words changes during training. We see that the word-level model is biased towards recognizing the most common words and slowly learns less frequent words over time. 
	This makes sense since more weight is given to the corresponding examples. 
	While the same effect is present in the character-level model, it covers the complete support of the word frequency distribution in the same number of steps.
	
	On the other hand for the length distribution, we see that the word-level model covers all words independent of its length within the beginning of training. 
	The character-level model focuses strongly on shorter words before it covers the whole range of the word length distribution. If we compare the learning dynamics of both models, we find that each model learns words with different characteristics more easily. 
	If we take a look at the MTL model, we see that it combines both biases and arrives at learning a distribution that is much more uniform across both word frequency and word length. 
	We hypothesize that putting more emphasis on the tail of each of these distributions combines the strengths of the two models and makes them perform better, especially in distributions that follow a power law such as word frequency rank.
	\section{Conclusion}
	In contrast to earlier studies in the literature, we found that, even on a relatively small dataset, training on a word-level can be feasible. 
	Furthermore, we found that combining a word-level model with character-level supervision in MTL can improve results noticeably. 
	To gain a better understanding of this, we characterized the inductive bias of word-character MTL in ASR by comparing the distributions of recognized words at the beginning of training. 
	We found that adding character-level supervision to a word-level interpolates between recognizing more frequent words (preferred by the word-level model) and shorter words (preferred by the character-level model). 
	This effect could be even more pronounced on harder datasets than WSJ, such as medical communication data where many long words are infrequent, but very important. 
	Further analysis of word distributions in terms of pitch, noise and acoustic variability could provide additional insight.
	\bibliography{bibliography}
	\bibliographystyle{abbrvnat}
	\newpage
	\appendix
	\section{Architecture}
	\label{sec:arch}
	The architecture we used throughout the paper is based on Wav2Letter \citep{collobert_wav2letter_2016}. It is shown in Table~\ref{tbl:architecture}. Different from the original setup we use 2D convolutions in the first layers following the input and a slightly larger network. We apply dropout, batch normalization and ReLU activations to every layer but the input and last layer. We use a dropout rate of $0.2$ for the convolutional layers and $0.4$ for the dense layers. We clip the ReLU activations at a value of $20$. Here, we show the architecture of the character-level model. The word-level model only differs in a larger output dimension ($9411$ instead of $32$). For training we add a CTC loss on top of the dense layer, and for inference a softmax output.
	\begin{table}[h]
		  \caption{Neural network architecture of the character-level model.}
		\label{tbl:architecture}
		\centering
		\begin{tabular}{rrrrr}
			\toprule
			\multicolumn{1}{c}{Layer} & \multicolumn{1}{c}{Dimensions} & \multicolumn{1}{c}{Kernel } & \multicolumn{1}{c}{Strides} & \multicolumn{1}{c}{Filters}  \\ 
						\multicolumn{1}{c}{} & \multicolumn{1}{c}{[time, freq., channel]} & \multicolumn{1}{c}{[time, freq.]} & \multicolumn{1}{c}{[time, freq.]} & \multicolumn{1}{c}{}  \\ 
			\midrule
				 input & $[2500, 40, 1]$ &  &  & \\
				conv2d & $[1250, 20, 64]$ & $[11, 15]$ & $[2, 2]$ & $64$ \\
				conv2d-1 & $[1250, 10, 64]$ & $[11, 7]$ & $[1, 2]$ & $64$ \\
				conv2d-2 & $[1250, 5, 192]$ & $[11, 7]$ & $[1, 2]$ & $192$ \\
				reshape-conv2d-to-conv1d & $[1250, 960]$ &  &  & \\
				conv1d & $[1250, 256]$ & $7$ & $1$ & $256$ \\
				conv1d-1 & $[1250, 256]$ & $7$ & $1$ & $256$ \\
				conv1d-2 & $[1250, 256]$ & $7$ & $1$ & $256$ \\
				conv1d-3 & $[1250, 256]$ & $7$ & $1$ & $256$ \\
				conv1d-4 & $[1250, 256]$ & $7$ & $1$ & $256$ \\
				conv1d-5 & $[1250, 256]$ & $7$ & $1$ & $256$ \\
				conv1d-6 & $[1250, 256]$ & $7$ & $1$ & $256$ \\
				conv1d-7 & $[1250, 2048]$ & $32$ & $1$ & $2048$ \\
				dense & $[1250, 2048]$ &  &  & \\
				dense-1 & $[1250, 32]$ &  &  & \\
			\bottomrule
		\end{tabular}
	\end{table}
	
\end{document}